\documentclass[journal,twoside,web]{ieeecolor}
\usepackage{generic}
\usepackage{cite}
\usepackage{amsmath,amssymb,amsfonts}
\usepackage{algorithmic}
\usepackage{graphicx}
\usepackage{textcomp}
\usepackage{enumerate}

\usepackage{array}
\usepackage{stfloats}
\usepackage{url}
\usepackage{verbatim}
\usepackage{graphicx}
\usepackage{subcaption}
\usepackage[ruled]{algorithm2e}
\usepackage{pgfplots}
\usepackage{diagbox}
\usepackage{booktabs}
\usepackage{multirow}
\usepackage{float}
\usepackage{bm}
\usepackage{hyperref}
\pgfplotsset{width=8cm,compat=1.9}
\usepgfplotslibrary{fillbetween}

\DeclareMathOperator*{\argmax}{arg\,max}
\DeclareMathOperator*{\milds}{\mathnormal{MI}\text{-}\mathnormal{LDS}}

\DeclareMathOperator*{\E}{\mathbb{E}}
\graphicspath{{./figs/}}

\def\BibTeX{{\rm B\kern-.05em{\sc i\kern-.025em b}\kern-.08em
    T\kern-.1667em\lower.7ex\hbox{E}\kern-.125emX}}
\begin{document}
\title{Leveraging unlabelled data in
	multiple-instance learning problems for improved detection of 
	Parkinsonian tremor in free-living conditions}
\author{Alexandros Papadopoulos, and Anastasios Delopoulos, \IEEEmembership{Member, IEEE}
  \thanks{ Submission date: \today}
  \thanks{ This work was supported by European Union’s Horizon 2020 research and
  innovation programme under grant agreement No 965231.} 
  \thanks{Alexandros
  Papadopoulos (alpapado@mug.ee.auth.gr) and Anastasios Delopoulos
  (antelopo@ece.auth.gr) are with the ECE Department of Aristotle University of
  Thessaloniki. } 
}

\maketitle

\begin{abstract}
    Data-driven approaches for remote detection of Parkinson's Disease and its
    motor symptoms have proliferated in recent years, owing to the potential
    clinical benefits of early diagnosis. The holy grail of such approaches is
    the free-living scenario, in which data are collected continuously and
    unobtrusively during every day life. However, obtaining fine-grained
    ground-truth and remaining unobtrusive is a contradiction and therefore, the
    problem is usually addressed via multiple-instance learning. Yet for large
    scale  studies, obtaining even the necessary coarse ground-truth is not
    trivial, as a complete neurological evaluation is required. In contrast,
    large scale collection of data without any ground-truth is much easier.
    Nevertheless, utilizing unlabelled data in a multiple-instance setting is
    not straightforward, as the topic has received very little research
    attention.  Here we try to fill this gap by introducing a new method for
    combining semi-supervised with multiple-instance learning. Our approach
    builds on the Virtual Adversarial Training principle, a state-of-the-art
    approach for regular semi-supervised learning, which we adapt and modify
    appropriately for the multiple-instance setting. We first establish the
    validity of the proposed approach through proof-of-concept experiments on
    synthetic problems generated from two well-known benchmark datasets.  We
    then move on to the actual task of detecting PD tremor from hand
    acceleration signals collected in-the-wild, but in the presence of
    additional completely unlabelled data. We show that by leveraging the
    unlabelled data of 454 subjects we can achieve large performance gains
    (up to 9\% increase in F1-score)  in per-subject tremor detection for a
    cohort of 45 subjects with known tremor ground-truth. In doing so, we
    confirm the validity of our approach on a real-world problem where the need
    for semi-supervised and multiple-instance learning arises naturally.
\end{abstract}


\section{Introduction}

\label{sec:introduction}
\IEEEPARstart{P}{arkinson's} disease (PD) is a long-term neurological
disorder associated with both motor and non-motor symptoms, like extremity tremor, 
bradykinesia, rigidity, degradation of fine-motor skills, depressive tendencies,
speech impairment and sleep abnormalities \cite{Jankovic368}. Despite being incurable for the time
being, the stage of the disease at first diagnosis can have a large impact on
the disease progression and the patient's quality of life, with early diagnosis
linked to significant improvements in both of these outcomes
\cite{pagan2012improving}. This link, along with the fact that early signs of PD
are often unnoticed or ignored by patients, has jumpstarted a line of research
\cite{mei2021machine} into using machine learning (ML) on sensor data for
continuously and objectively monitoring an individual for patterns or
behavioural changes that might indicate PD.

To this end, various sensor types and algorithms have been proposed, targetting
different symptoms of the disease. For instance, 
tapping on virtual or physical keyboards has been used for quantifying
fine-motor impairment \cite{keyboard_original, iakov, tripathi2022keystroke},
microphones for estimating the severity of speech degradation 
\cite{orozco2016towards, 9556632, deep_speech} and wearable accelerometers and
gyroscopes for detecting tremor\cite{s20205817, s100302129,
papadopoulos2019multiple},  gait imbalance\cite{imu_gait, gait2, gait3}  and
eating difficulties caused by bradykinesia \cite{kyritsis2021assessment}. 
Building on the basic idea of associating data from a specific type of sensor 
to a specific motor symptom, some works even follow a more holistic approach and
combine multiple sensors to detect PD itself and not just any one 
symptom \cite{papadopoulos2020unobtrusive, multi_modal1, multi_modal2,
multi_modal3}.

Most approaches in the literature, collect data in a controlled lab-based or, at
best, a scripted home-based environment. Few works actually operate in the
free-living setup, despite being the most suitable for large scale adoption of
any proposed technologies. The rather slow transition to the unobtrusive and
in-the-wild setting is not without reason, as the latter poses significant
difficulties to machine learning approaches, ranging from low signal-to-noise
ratios to difficulties in obtaining precise ground-truth for training and
evaluation. These difficulties are usually circumvented by using coarse,
subject-level annotations that allow predictions to be made using
multiple-instance learning  \cite{alpapado2019tremor, papadopoulos2019multiple,
alpapado2020pd, das2012detecting}. However, acquiring even a coarse annotation
can be challenging when collecting data from hundreds of people, and each one
must be subjected to a neurological evaluation. On the contrary, as we will
see, collecting unlabeled data from hundreds of subjects is much easier and
cheaper to accomplish. 

In this paper, we study the problem of predicting whether a person suffers from
PD-induced tremor by analyzing hand acceleration signals that
have been captured unobtrusively throughout  the daily user-smartphone
interaction. To this end, a data collection app that ``listens'' to the phone's
IMU sensor whenever a phone call is placed was employed (developed within the
context of \cite{lisa_klingelhoefer_2017_1199554}). 
Remote screening for PD tremor under
unobtrusive and free-living conditions was first tackled in
\cite{alpapado2019tremor} using a relatively small dataset of 45 subjects that
were all annotated for tremor at the subject level. Here we examine  
whether tremor detection performance can be
improved via the incorporation of an additional large dataset of 454 subjects,
collected under the same conditions, but lacking tremor annotation. This is a
natural next step, as acquiring additional unlabelled data is very easy: we just
need to  distribute the data collection app to as many people as possible. On
the contrary, acquiring labelled data requires the person to be
neurologically evaluated, a process both time-consuming and expensive
and thus not scalable to large numbers of participants.

Building ML models using few labelled and much unlabelled data corresponds to a
learning approach called \emph{semi-supervised learning}. Semi-supervised
learning (SSL) has received much attention from the research community over the
years, owing to the ease of obtaining vast amounts of data and to the inherent
difficulty in labelling them. In recent years, semi-supervised
approaches for deep learning have 
led to impressive results. 
Approaches like self-supervised learning \cite{9086055} and consistency-based
regularization \cite{46794} before it, have been successfully applied to many
image  classification problems, achieving performance comparable to their
fully-supervised counterparts, while using only a  handful of labels
\cite{verma2019interpolation, berthelot2019mixmatch, bachman2019learning}.

Currently, most SSL approaches work in the single-instance
learning setting, where the goal is to predict the label $y$ of a data
sample $\mathbf{x}$. However, a different setting with particular interest for
remote disease screening  is that of Multiple-Instance Learning (MIL), where the
goal is to predict a label $y$ for a bag of samples (or \emph{instances})
$\{\mathbf{x}_1, \mathbf{x}_2, \dots\}$. Throughout the learning procedure, the
labels of the instances in the bag are unknown and the  only available
annotation is a label that describes the entire bag. This situation is regularly
encountered in practice. In our case, for example, PD tremor may manifest only for a
fraction of time during everyday life, depending on the disease's stage, the
symptom's intermittence or levodopa intake.  Hence, to detect tremor from
sensor measurements obtained in-the-wild, we must resort to MIL, as there is no
easy way to obtain detailed ground-truth of the on-off periods.

Here we are interested in the combination of semi-supervised with
multiple-instance learning. In particular,
we are interested in whether we can use unlabelled bags
to improve a multiple-instance classifier.
Interestingly, this problem has received very little research attention.
The early work of \cite{rahmani2006missl} uses unlabelled bags
in a content-based image retrieval setting and proposes a way 
of transforming the MI problem to
a single-instance graph-based label propagation \cite{zhou2004learning} problem
that has the MI constraints encoded in the graph structure.
The subsequent work of \cite{tang2008integrated} uses a similar
graph-based approach and suggests a unification between
the representation of the image on the instance level
and on the bag level to transductively annotate images.
A further modification of the graph optimization objective
is proposed in \cite{jia2008instance}. An interesting approach is
the one by \cite{xu2012semi} who propose
a regularization-based MI SSL approach for video annotation tasks,
in which similar instances are encouraged to share similar labels,
while an instance-level label propagation scheme that combines
label propagation with MI is proposed in \cite{ijcai2017-410}.
Finally, \cite{10.1145/3447548.3467318} explores a scenario where a
MIL classifier is trained using similarity information between
bags, rather than complete labels. 

Based on the above overview, one notices that almost all of these works
are concerned with transductive SSL \cite{vapnik200624}, where
the goal is to assign labels to the given unlabelled data (i.e. propagate
labels to the bag instances), rather than learning a general mapping from the
data to the label domain. In addition, most related methods predate deep
learning and are focused on more traditional machine learning models. In this paper we
follow a different direction and propose an approach based on
consistency-regularization that is able to  leverage unlabelled bags in order to
improve a deep multiple-instance learning classifier. Concretely, we make the
following contributions:
\begin{enumerate}[(i)]
        \item We propose a method for leveraging unlabelled bags of instances in
          multiple-instance problems, by adapting a state-of-the-art
          semi-supervised algorithm to the multiple-instance setting.
          Furthermore, we introduce two additional modifications to the main
          algorithm that further boost performance. 
        \item To demonstrate its validity in a controlled environment, we 
          conduct thorough experiments in synthetically-generated datasets based
          on the MNIST and CIFAR-10 datasets, where we demonstrate systematic
          improvements in performance from the incorporation of unlabelled data,
          compared to the fully supervised baseline and two alternative
          state-of-the-art SSL algorithms.
        \item We introduce a new dataset of hand acceleration recordings
          captured unobtrusively from PD patients and Healthy controls during
          free-living conditions. It consists of a small cohort of 45 labelled
          (with  tremor ground-truth) and a large cohort of 454 unlabelled
          (without tremor ground-truth) subjects. It serves as an extension of a
          previous dataset  and is made publicly available
          \href{https://doi.org/10.5281/zenodo.7273759}{here}.
        \item By utilizing this unlabelled cohort, we demonstrate that our
          proposed approach can lead  to large and systematic improvements in PD tremor
          detection performance (up to $\sim$9\% increase in F1-score) compared to the
          alternative  that makes use of just the small labelled cohort,
          thus demonstrating its utility in remote disease screening.
\end{enumerate}

The rest of this paper is organized as follows. In Section \ref{sec:prelim} we
present the related literature for the SSL and MIL domains, focusing mostly on
approaches that are related to ours. In Section \ref{sec:approach} we
introduce our method for MI SSL and its proposed variants. Section
\ref{sec:experiments} presents the initial, proof-of-concept experimental
results. Then section \ref{sec:tremor_exp}
introduces the in-the-wild tremor detection problem and demonstrates how our
approach can be applied to improve performance in this problem. Finally, in
\ref{sec:discussion} we discuss the potential benefits and caveats of the
presented approach.

\section{Preliminaries}

\label{sec:prelim}

\subsection{Semi-Supervised Learning}

Semi-supervised learning (SSL) is a situation where in addition to a
fully labelled set $\mathcal{D}_l = \{ (\mathbf{x}_i, y_i) \}_{i=1}^L$, we are also
presented with a set of unlabelled data points $\mathcal{D}_u = \{
\mathbf{x}_i \}_{i=L+1}^{L+U}$ drawn i.i.d. from the same marginal distribution. 
The goal is to leverage $\mathcal{D}_u$ in order to learn a more accurate classifier
than what would be possible using only $\mathcal{D}_l$. In general, it is not evident how
unlabelled data can help, as knowledge of the
marginal  does not directly contribute to the data likelihood for a given model.
In fact, $\mathcal{D}_u$ can be helpful only if certain assumptions are true.
The most common one is
the \emph{smoothness assumption}, which states that if two points
$\mathbf{x}_1$, $\mathbf{x}_2$ in a high-density  region are close, then so
should be their predicted labels $y_1$, $y_2$. This suggests that the learnt
classifier must be smooth in high-density regions. The well-known
\emph{low-density separation} assumption that requires the decision boundary
to lie in a low-density region, is an alternative view of the smoothness assumption.

Early SSL techniques for neural networks were designed to enforce the low-density
separation assumption on the resulting classifier. They did so by penalizing
a decision boundary in high-density regions, for example by
encouraging the model output distribution to have low entropy \cite{ssl_book}.
More recent techniques employ a similar regularization scheme, in which the model
is encouraged to be invariant across label-preserving transformations
of the input data (e.g. a small amount of additive gaussian noise that does not
change the label).
This principle is called \emph{consistency regularization} and is used in many
state-of-the-art methods for semi-supervised image classification, like
Pseudo-Ensembles \cite{sajjadi2016regularization},  Temporal Ensembling
\cite{laine2016temporal} and  Mean Teacher \cite{tarvainen2017mean}.

One approach to consistency regularization that is of particular interest in this work,
is the \emph{Virtual Adversarial Training (VAT)} method \cite{Miyato2016a}.
Instead of perturbing $\mathbf{x}$ randomly, VAT calculates
the perturbation that will cause the 
maximum change in the model's output. 
To achieve this, at each training step, it solves
the following optimization problem:
\begin{equation}
\mathbf{r}_{vadv} = \argmax_{\mathbf{r} ; ||\mathbf{r}||_2 = \epsilon} 
D\left[p(y|\mathbf{x}; \hat{\theta}), p(y|\mathbf{x+r}; \hat{\theta})\right]
\label{eq:vat}
\end{equation}
where $p(y|\mathbf{x}; \hat{\theta})$ is our model, $D$ is 
a distribution divergence metric and $\hat{\theta}$ denotes the
model parameters at the current step. 
The optimization problem of Eq. \ref{eq:vat} 
can be approximated efficiently
with just an additional forward-backward pass through the network.
Having estimated $\mathbf{r}_{vadv}$,
the model is then encouraged to be smooth along its direction by minimizing the
\emph{Local Distributional Smoothing (LDS)} loss at each data point:
\begin{equation}
LDS = D\left[p(y|\mathbf{x}; \hat{\theta}), p(y|\mathbf{x}+\mathbf{r}_{vadv}; \hat{\theta})\right]
\end{equation}
Empirical results suggest that encouraging consistency 
along the virtual adversarial
direction, $\mathbf{r}_{vadv}$, results in 
significantly improved performance compared
to consistency in random perturbations.
In the following, we will present an approach 
for semi-supervised multiple-instance learning
that is based on VAT. We elect this particular technique because its
ultimate goal is conceptually intuitive and elegant and its background is
theoretically sound.

\subsection{Multiple-Instance Learning}
In Multiple-Instance Learning (MIL) we are again presented with a set of samples
and their labels $D = \{ (X_i, y_i) \}_{i=1}^L$. The difference is that here
each sample is itself a bag instances, i.e.
$X_i=\{\mathbf{x}_i^1, \mathbf{x}_i^2, \dots, \mathbf{x}_i^K\}$  with
$\mathbf{x}_i^j \in \mathbb{R}^N$, while $y_i$ refers to the entire bag
$X_i$ and not to any one  instance $\mathbf{x}_i^j$. 
The goal in this scenario is to learn a bag classifier.
Since a bag is an unordered set of instances without dependencies
between its members, our classifier should be
permutation-invariant
with respect to the ordering of the bag instances.
Theoretical results \cite{zaheer2017} suggest that 
a bag function $f(X)$ is permutation-invariant if and only if
it can be decomposed in the form:
\begin{equation}
f(X) = \rho\left(\mathbf{z}\right),  \quad \mathbf{z} = \sum_{\mathbf{x} \in X} \phi(\mathbf{x})
\label{eq:sum_decomp}
\end{equation}
where $\phi$ is an embedding function $\mathbb{R}^N \mapsto \mathbb{R}^M$,
$\mathbf{z} \in \mathbb{R}^M$ is the embedding of $X$ and
$\rho$ a suitable transformation $\mathbb{R}^M \mapsto \mathcal{Y}$ (with
$\mathcal{Y}$ we denote the label domain).

For models based on neural networks,  the transformation $\phi$ is usually a
high-capacity CNN that is used either for feature extraction or for direct
instance classification. The transformation $\rho$ is then either a
classification head that takes us  from the embedding space to the class space,
or simply  the identity.
%
A rather interesting modification to the above stems from incorporating an attention
mechanism on the sum of Equation \ref{eq:sum_decomp}.
This approach \cite{pmlr-v80-ilse18a} defines the bag embedding as  a non-linear
combination (with learnable parameters $\mathbf{V}, \mathbf{w}$)  of the instance features:
\begin{equation}
\mathbf{z} = \sum_{k=1}^K \alpha_k \phi(\mathbf{x}_k),
\quad \alpha_k = \frac{e^{\mathbf{w}^T tanh(\mathbf{V}\phi(\mathbf{x}_k)^T)}}{\sum_{i=1}^K
\; e^{\mathbf{w}^T tanh(\mathbf{V}\phi(\mathbf{x}_i)^T)}}\label{eq:attention_mil}
\end{equation}

The attention parameters $\mathbf{V}$ and $\mathbf{w}$  can be easily modelled
as neural networks, thus  allowing the whole model to be learnable end-to-end. 
In addition, attention scores provide an elegant way of identifying key
instances within a bag.  Attention-based MIL has been successfully applied to
many problems \cite{alpapado2020pd,9098062,alpapado2021dr}.  Owing to its
attractive properties and performance, we will use it as our core MIL model,
which we will enhance in the next section with a semi-supervised
component. 

\section{Semi-supervised multiple-instance learning}
\label{sec:approach}
In this section, we present our approach for utilizing unlabelled bags of
instances in order to improve a multiple-instance classifier.  As VAT provides a
principled and elegant way of using unlabelled data, we elect to use it for
semi-supervised MIL, over other SSL approaches (e.g. Mean Teachers) 
that could be directly applied to
the same problem. 

First, we extend VAT to the multiple-instance scenario.
To that end, we introduce the concept of bag perturbation 
which is a set $R = (\mathbf{r}_1, \mathbf{r}_2, \dots, \mathbf{r}_K)$
that when added elementwise to a given bag $X$, slightly perturbs it.
The Multiple-Instance Local Distributional Smoothing (MI-LDS) loss can now be defined as:
\begin{align}
	\milds(X, \hat{\theta}) &= D\left[p(y|X;\hat{\theta}), p(y|X+R\textsubscript{vadv};\hat{\theta})\right]
	\label{eq:chap8:lds_loss}
\end{align}
where 
\begin{equation*}
  \quad
\begin{split}
  &X = (\mathbf{x}_1, \mathbf{x}_2, \dots \mathbf{x}_K) \\
    &R\textsubscript{vadv} = \left(\mathbf{r}_1, \mathbf{r}_2, \dots,
        \mathbf{r}_K\right) \\
  &\qquad\;= \argmax_{R; ||r_k||_2 <= \epsilon} D[p(y|X; \hat{\theta}), p(y|X+R; \hat{\theta})]\\
               &X+R\textsubscript{vadv} = \left(\mathbf{x}_1 + \mathbf{r}_1, \mathbf{x}_2 + \mathbf{r}_2, \dots, \mathbf{x}_K + \mathbf{r}_K\right)
\end{split}
\end{equation*}
and $D$ is a distribution divergence like KL divergence.
As MI-LDS does not depend on the bag label, it can be added as an unsupervised regularization
term in the loss function. 
Assuming a set of labelled bags $\mathcal{D}_l = \{(X_l^i, y_l^i)\} |_{i=1}^L$ and a
set of unlabelled bags $\mathcal{D}_{ul} = \{X_{ul}^j\} |_{j=1}^U$, the loss
function is
\begin{equation}
  \mathcal{L} = \E_{X,y \sim \mathcal{D}_l} \left[ -\log p(y|X; \hat{\theta})
\right]
+ \E_{X \sim \mathcal{D}_{ul}} \left[\milds(X, \hat{\theta})\right]
\label{eq:chap8:cost_function}
\end{equation}
where the first term is the standard cross-entropy loss computed over labelled
bags and the second term is the MI-LDS loss computed over the unlabelled bags.


We will now show how to compute $R\textsubscript{vadv}$ based on the original
approximation method of the virtual perturbation direction in the
single-instance scenario\cite{Miyato2016a}.
Without loss of generality, we will treat $R$ as a flattened vector of the form
$[r_1^1, \dots, r_{N}^1, r_{1}^2, \dots r_{N}^2, \dots, r_{1}^K, \dots r_{N}^K]$.
For brevity, we denote 
$D\left[p(y|X;\hat{\theta}), p(y|X+R\textsubscript{vadv};\hat{\theta})\right]$ by $D\left(R,X,\hat{\theta}\right)$.
We also assume that $p(y|X; \hat{\theta})$ is differentiable twice with respect to $\hat{\theta}$ and $X$
almost everywhere. Since $D\left(R,X,\hat{\theta}\right)$ is minimized at 
$R_0=\left(\mathbf{0}, \mathbf{0}, \dots, \mathbf{0}\right)$ and given the differentiability
assumption, we have that 
$\nabla_R D(R, X; \hat{\theta}) |_{R=R_0} = \mathbf{0}$.
Hence, its second-order Taylor approximation at $R=R_0$ is given by
\begin{equation}
D\left(R,X,\hat{\theta}\right) \approx \frac{1}{2} R^T H(X; \hat{\theta}) R
\end{equation}
where $H(X; \hat{\theta}) := \nabla\nabla_R D\left(R,X,\hat{\theta}\right) |_{R=R_0}$ 
is the Hessian matrix. Under this approximation,
computing $R\textsubscript{vadv}$ reduces to computing the first dominant eigenvector of the Hessian.
This can be achieved by means of the power-iteration approach: we simply start out
with a randomly sampled perturbation $V$ and compute the following product
\begin{equation}
d \gets \overline{HV}
\end{equation}
where the overline operator $\overline{d}$ denotes normalization 
of each element of $d$ (which here is a bag entity) to unit length.
Repeating this calculation iteratively makes $d$
converge to the first dominant eigenvector of $H$.
Interestingly enough, the $HV$ product can be approximated 
via a finite difference method, thus eliminating the need
for directly computing $H$:
\begin{align}
  HV &\approx \frac{\nabla_R D(R, X, \hat{\theta})|_{R=\xi V} - \nabla_R D(R, X,
	\hat{\theta})|_{R=R_0}}{\xi} \\
     &= \frac{\nabla_R D(R, X, \hat{\theta})|_{R=\xi V}}{\xi}
\end{align}
where $\xi$ is a small constant.
Putting it all together, we can approximate 
$R\textsubscript{vadv}$ with the repeated calculation of
\begin{equation}
V \gets \overline{\nabla_R D\left(R,X,\hat{\theta}\right) |_{R=\xi \cdot V}} \label{eq:chap8:power_iter2}
\end{equation}
where initially $V$ is a bag of randomly sampled unit vectors.
In each iteration of Equation \ref{eq:chap8:power_iter2} we first compute the gradient of $D$ on
$R=\xi V$ and then normalize each element to unit length.
Based on the above, the virtual adversarial bag perturbation
$R\textsubscript{vadv}$ can be approximated using the one-time power iteration
approach of Algorithm \ref{algo:vadv}.

\begin{figure*}[!h]
  \centering
      \begin{subfigure}[b]{\textwidth}
       \centering
       \includegraphics[width=\textwidth]{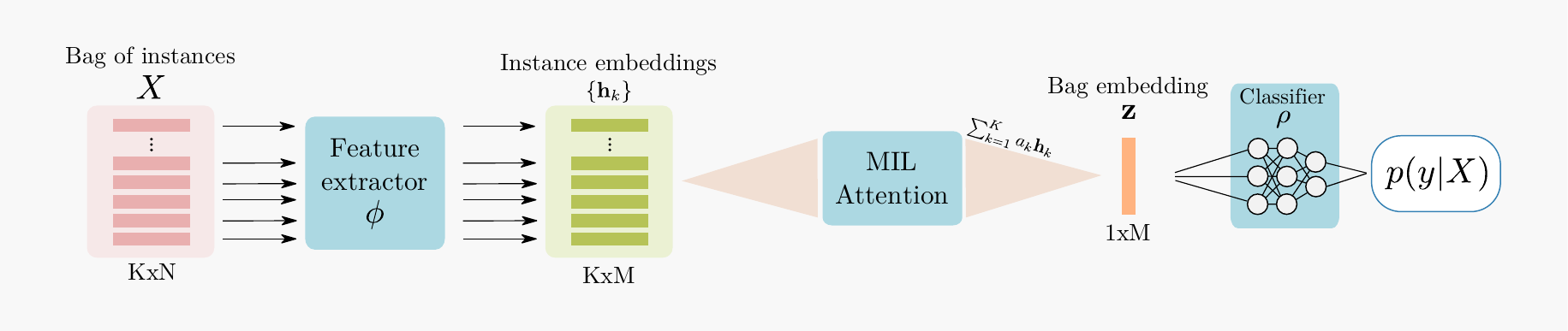}
       \caption{The attention-based deep MIL model.}
       \label{fig:mil_attention}
     \end{subfigure}
      \begin{subfigure}[b]{\textwidth}
       \centering
       \includegraphics[width=\textwidth]{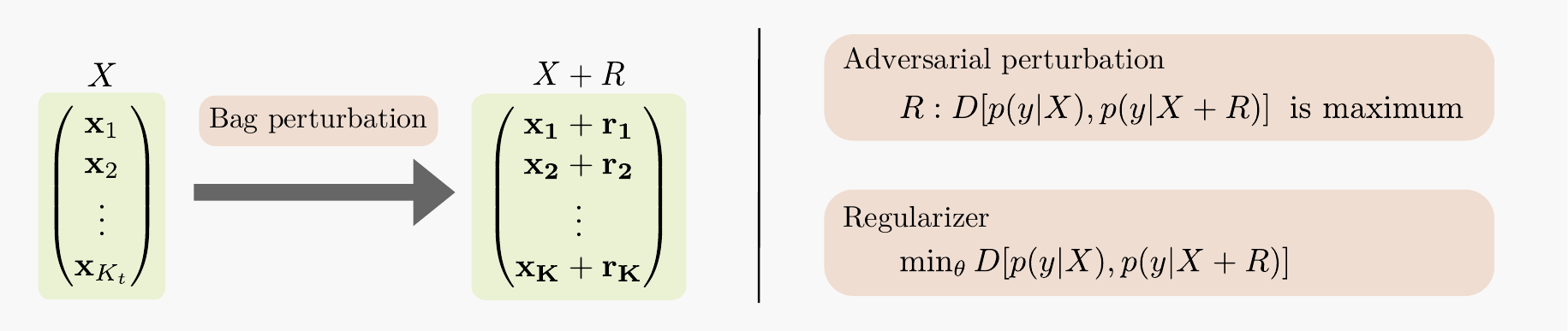}
       \caption{The bag perturbation idea in a nutshell.}
       \label{fig:y equals x}
     \end{subfigure}
     \caption{High-level overview of the main components involved in the
     proposed semi-supervised MIL methodology.}
        \label{fig:method_overview}
\end{figure*}

\begin{algorithm}[]
	\SetKwInOut{Input}{Input}\SetKwInOut{Output}{Output}
	\SetAlgoLined
	\Input{Bag $X = (\mathbf{x}_1, \mathbf{x}_2, \dots
		\mathbf{x}_K)$, model $p(y|X;\theta)$}
	\Output{Virtual adversarial bag perturbation $R\textsubscript{vadv}$}
	1) Generate a random bag $V = (\mathbf{v}_1, \mathbf{v}_2, \dots \mathbf{v}_K)$ 
	with $\mathbf{v}_k \sim N(\mathbf{0}, I)$\\
	
	2) Compute $p(y|X; \hat{\theta})$ and $p(y|X+\xi V; \hat{\theta})$
	
	3) Compute the (Kullback-Leibler) divergence 
	\begin{equation*}
		L = D_{KL}\left(p(y|X;\hat{\theta}) \parallel p(y|X+R;\hat{\theta})\right)
	\end{equation*}
	
	4) Calculate the gradient of $L$ on $R=\xi V$:
	\begin{equation*}
		G = (\mathbf{g}_1, \mathbf{g}_2, \dots, \mathbf{g}_K) \gets \nabla_R L |_{R=\xi \cdot V}
	\end{equation*}
	
	5) Normalize elementwise to the desired magnitude.
	\begin{equation*}
		R\textsubscript{vadv} \gets \left(\epsilon\frac{{\mathbf{g}}_1}{||\mathbf{g}_1||_2}, \epsilon\frac{\mathbf{g}_2}{||\mathbf{g}_2||_2}, \dots, \epsilon\frac{\mathbf{g}_K}{||\mathbf{g}_K||_2}\right)
	\end{equation*}
	
	\caption{Approximation of $R\textsubscript{vadv}$ with a one-time power iteration approach}
	\label{algo:vadv}
\end{algorithm} 


This procedure results in a bag perturbation that is dense, meaning
that all instances in the input bag $X$ are going to be perturbed:

\begin{center}
\textbf{Variant 1 - Regular (dense) perturbation}
	\begin{gather*}
		R\textsubscript{vadv} = \left(\smash[b]{\underbrace{\mathbf{r}_1, \mathbf{r}_2,
				\dots, \mathbf{r}_{K}}_{\text{\footnotesize K non-zero instances}}}\right)
	\end{gather*}
\end{center}

\vspace{3mm}

An interesting variation of this approach would be to make 
$R\textsubscript{vadv}$ sparse, i.e. restrict it so that it has a single
non-zero instance.
The hope in doing so is that by perturbing one bag instance instead of all, 
we will arrive at more meaningful bag perturbations overall. This is
due to the fact that in MIL,
changing the (unobserved) label of a single 
instance can change the label
of the bag itself, as a positive bag can contain as little as a single
positive instance (in the case of the standard MI assumption \cite{foulds2010review}).
Therefore, in such cases it may prove more beneficial
to perturb the few positive instances
of a positive bag, rather than all of its instances indiscriminately.
In practice, however, we do not know which are the positive instances
in a bag in order to perturb them in a targeted manner. Thus, to 
mitigate this obstacle we opt for a stochastic approach, in which
we select the index $j$ of the instance to be perturbed at random.

We propose two different strategies for choosing $j$. 
The first approach is to simply select $j$ uniformly at random over the $K$
possible instances. This approach leads to the following perturbation variant:

\begin{center}
  \textbf{Variant 2 - Sparse perturbation with uniform probability}
	\begin{gather*}
		R\textsubscript{vadv} = \left(\smash[b]{\underbrace{\mathbf{0}, \dots \mathbf{0}, \;\mathbf{r}_j, \;\mathbf{0}, \dots \mathbf{0}}_{\text{\footnotesize single non-zero instance}}} \right) \\\\
		\text{where} \quad p(j=k) = 1/K, \;\;\; 1 \leq k \leq K
	\end{gather*}
\end{center}

Let us now assume that $p(y|X;\hat{\theta})$ is the attention-based MIL model defined in Equations \ref{eq:sum_decomp} - \ref{eq:attention_mil}, that is:
\begin{equation}
p\left(y | X; \hat{\theta}\right) = 
\rho\left(\sum_k \alpha_k \phi\left(\mathbf{x}_k\right) \right)
\label{eq:class_score}
\end{equation}
where $\phi, \rho, \alpha_k$ are learnable. 
Given this model, a third variant for the
perturbation form is to sample $j$
from a multinomial distribution whose parameters are given by the attention
weights $\alpha_k$, as
$\{\alpha_k\}$ defines a multinomial distribution over $K$
outcomes since $0 \leq \alpha_k
\leq 1$ and $\sum_k \alpha_k = 1$. This can be written concisely as:

\begin{center}
  \textbf{Variant 3 - Sparse perturbation based on attention}
	\begin{gather*}
		R\textsubscript{vadv} = \left(\smash[b]{\underbrace{\mathbf{0}, \dots \mathbf{0}, \;\mathbf{r}_j, \;\mathbf{0}, \dots \mathbf{0}}_{\text{\footnotesize single non-zero instance}}} \right) \\\\
		\text{where} \quad p(j=k) = \alpha_k \;\; (\text{per
                  Eq. \ref{eq:attention_mil})}, \;\;\; 1 \leq k \leq K
	\end{gather*}
\end{center}

In this way, the non-zero index $j$ is sampled
from a different distribution for each bag, with instances that
receive high attention (which hopefully
correspond to key instances) more likely to be selected
in each training step.

In the following, we will explicitly use the attention-based model as our underlying
$p(y|X; \hat{\theta})$, because, apart from its attractive properties and performance, it allows
us to seamlessly accommodate variants 2 and 3 in Algorithm \ref{algo:vadv} by
means of a simple masking system. 
A high-level overview of the components involved in the proposed
methodology is presented in Fig. \ref{fig:method_overview}.

\section{Proof-of-concept experiments}
\label{sec:experiments}
Prior to addressing tremor detection in-the-wild, we wish to establish the
validity of our approach in a more controlled environment. To that end, in this
section we conduct some initial proof-of-concept experiments on two
semi-supervised  multiple-instance problems, that are synthetically generated
from two standard image classification datasets:
\begin{itemize}
  \item MNIST \footnote{\url{http://yann.lecun.com/exdb/mnist/}}, a 
    dataset of 70,000 28x28 grayscale images of handwritten digits (0
    to 9), split in a training set of 60,000 and a test set 10,000 images.

  \item CIFAR-10 \footnote{\url{https://www.cs.toronto.edu/~kriz/cifar.html}}, a
    dataset containing 60,000 32x32 color images of the following
    mutually exclusive classes: airplane, automobile, bird, cat, deer, dog,
    frog, horse, ship, truck. Its training/test split is 50,000/10,000.

\end{itemize}

To create MIL problems from these datasets we will define a bag as a set of
random images. For MNIST, a bag will be positive if it  contains at least one
image of the digit "9" (as per \cite{pmlr-v80-ilse18a}), while for CIFAR-10, a
positive bag will have to contain at least one image from the "Truck" class. The
length $K$ of each bag is randomly sampled from the Gaussian distribution
$\mathcal{N}(K_\textsubscript{mean}, K_\textsubscript{std})$.  The number of
positive samples in a positive bag is sampled from the uniform
distribution $U(1, K)$, while class imbalance is controlled by a parameter
$p_1$. 

For our experiments, we create a training set of $L$ labelled and $U$ unlabelled
bags from the official training set of each dataset. The instances in each bag
are sampled without replacement, so that bags do not contain common instances.
For testing, we sample a set of $1.000$ bags from the official test set using
the same parameters. For efficiency, we opt for relatively small bag sizes, thus
setting $K_\textsubscript{mean}$=10 and $K_\textsubscript{std}$=2 in the data
generation process (larger bag sizes arise naturally in the context of our
tremor detection problem of Section \ref{sec:tremor_exp}). In addition, positive
class imbalance was set to 1 positive for every 10 negative ($p_1$=0.1)  in
order to simulate the case of highly imbalanced classes that is encountered in
real-world datasets (see Section \ref{sec:tremor_exp}). 

\begin{table}[!ht]
	\centering
	\caption{Average ROC-AUC scores across 10 trials of the MNIST experiment. 
		Positive bags are those with at least one "9" digit.}
        \resizebox{\columnwidth}{!}{%
	\begin{tabular}{c|*{3}{c}}
		\toprule
		Labelled 
		& Unlabelled 
		& Method
		& ROC-AUC \\
		\midrule
		\multirow{11}{*}{\parbox{1cm}{\centering 50}} 
		& \multirow{1}{*}{\parbox{1cm}{\centering 0}}
		& \multirow{1}{*}{\parbox{1cm}{\centering Baseline}}
		& \multirow{1}{*}{\parbox{2cm}{\centering 0.702 $\pm$ 0.057}} 
		\\\cmidrule{2-4}
		
		& \multirow{5}{*}{\parbox{1cm}{\centering 200}}
		& \multirow{1}{*}{\parbox{3.1cm}{\centering Temporal ensembling\cite{laine2016temporal}}}
		& \multirow{1}{*}{\parbox{2cm}{\centering 0.768 $\pm$ 0.019}} 
		\\
		&
		& \multirow{1}{*}{\parbox{3cm}{\centering Mean teacher\cite{tarvainen2017mean}}}
		& \multirow{1}{*}{\parbox{2cm}{\centering 0.833 $\pm$ 0.025}} 
		\\
		&
		& \multirow{1}{*}{\parbox{3cm}{\centering Dense MI-VAT}}
		& \multirow{1}{*}{\parbox{2cm}{\centering 0.859 $\pm$ 0.043}} 
		\\
		
		& 
		& \multirow{1}{*}{\parbox{3cm}{\centering Sparse-Uniform MI-VAT}}
		& \multirow{1}{*}{\parbox{2cm}{\centering 0.927 $\pm$ 0.020}} 
		\\
		
		& 
		& \multirow{1}{*}{\parbox{3cm}{\centering Sparse-Attention MI-VAT}}
		& \multirow{1}{*}{\parbox{2cm}{\centering \textbf{0.943} $\pm$ 0.025}} 
		\\
		\cmidrule{2-4}
		& \multirow{5}{*}{\parbox{1cm}{\centering 400}}
		& \multirow{1}{*}{\parbox{3.1cm}{\centering Temporal ensembling\cite{laine2016temporal}}}
		& \multirow{1}{*}{\parbox{2cm}{\centering 0.734 $\pm$ 0.048}} 
		\\
		&
		& \multirow{1}{*}{\parbox{3cm}{\centering Mean teacher\cite{tarvainen2017mean}}}
		& \multirow{1}{*}{\parbox{2cm}{\centering 0.903 $\pm$ 0.012}} 
		\\
		&
		& \multirow{1}{*}{\parbox{3cm}{\centering Dense MI-VAT}}
		& \multirow{1}{*}{\parbox{2cm}{\centering 0.881 $\pm$ 0.040}} 
		\\
		
		& 
		& \multirow{1}{*}{\parbox{3cm}{\centering Sparse-Uniform MI-VAT}}
		& \multirow{1}{*}{\parbox{2cm}{\centering 0.918 $\pm$ 0.029}} 
		\\
		
		& 
		& \multirow{1}{*}{\parbox{3cm}{\centering Sparse-Attention MI-VAT}}
		& \multirow{1}{*}{\parbox{2cm}{\centering \textbf{0.935} $\pm$ 0.033}} 
		\\
		\midrule[0.7pt]
		\multirow{11}{*}{\parbox{1cm}{\centering 100}} 
		& \multirow{1}{*}{\parbox{1cm}{\centering 0}}
		& \multirow{1}{*}{\parbox{2cm}{\centering Baseline}}
		& \multirow{1}{*}{\parbox{2cm}{\centering 0.811 $\pm$ 0.052}} 
		\\\cmidrule{2-4}
		
		& \multirow{5}{*}{\parbox{1cm}{\centering 200}}
		& \multirow{1}{*}{\parbox{3.1cm}{\centering Temporal ensembling\cite{laine2016temporal}}}
		& \multirow{1}{*}{\parbox{2cm}{\centering 0.904 $\pm$ 0.026}} 
		\\
		&
		& \multirow{1}{*}{\parbox{3cm}{\centering Mean teacher\cite{tarvainen2017mean}}}
		& \multirow{1}{*}{\parbox{2cm}{\centering 0.926 $\pm$ 0.010}} 
		\\
		&
		& \multirow{1}{*}{\parbox{3cm}{\centering Dense MI-VAT}}
		& \multirow{1}{*}{\parbox{2cm}{\centering 0.973 $\pm$ 0.026}} 
		\\
		
		& 
		& \multirow{1}{*}{\parbox{3cm}{\centering Sparse-Uniform MI-VAT}}
		& \multirow{1}{*}{\parbox{2cm}{\centering \textbf{0.986} $\pm$ 0.008}} 
		\\
		
		& 
		& \multirow{1}{*}{\parbox{3cm}{\centering Sparse-Attention MI-VAT}}
		& \multirow{1}{*}{\parbox{2cm}{\centering 0.984 $\pm$ 0.009}} 
		\\
		\cmidrule{2-4}
		& \multirow{5}{*}{\parbox{1cm}{\centering 400}}
		& \multirow{1}{*}{\parbox{3.1cm}{\centering Temporal ensembling\cite{laine2016temporal}}}
		& \multirow{1}{*}{\parbox{2cm}{\centering 0.872 $\pm$ 0.020}} 
		\\
		&
		& \multirow{1}{*}{\parbox{3cm}{\centering Mean teacher\cite{tarvainen2017mean}}}
		& \multirow{1}{*}{\parbox{2cm}{\centering 0.947 $\pm$ 0.019}} 
		\\
		&
		& \multirow{1}{*}{\parbox{3cm}{\centering Dense MI-VAT}}
		& \multirow{1}{*}{\parbox{2cm}{\centering \textbf{0.988} $\pm$ 0.009}} 
		\\
		
		& 
		& \multirow{1}{*}{\parbox{3cm}{\centering Sparse-Uniform MI-VAT}}
		& \multirow{1}{*}{\parbox{2cm}{\centering 0.980 $\pm$ 0.012}} 
		\\
		
		& 
		& \multirow{1}{*}{\parbox{3cm}{\centering Sparse-Attention MI-VAT}}
		& \multirow{1}{*}{\parbox{2cm}{\centering 0.982 $\pm$ 0.011}} 
		\\
		\bottomrule
	\end{tabular}
      }
	\label{tab:results_mnist2}
\end{table}

\begin{table}[!h]
	\centering
	\caption{Average AUC scores across 5 trials of the CIFAR-10 experiment.
		Positive bags are those with at least one image of the "Truck" class.}
        \resizebox{\columnwidth}{!}{%
	\begin{tabular}{c|*{3}{c}}
		\toprule
		Labelled
		& Unlabelled
		& Method
		& ROC-AUC \\
		\midrule
		\multirow{11}{*}{\parbox{1cm}{\centering 200}} 
		& \multirow{1}{*}{\parbox{1cm}{\centering 0}}
		& \multirow{1}{*}{\parbox{2cm}{\centering Baseline}}
		& \multirow{1}{*}{\parbox{2cm}{\centering 0.732 $\pm$ 0.067}} 
		\\\cmidrule{2-4}
		
		& \multirow{5}{*}{\parbox{1cm}{\centering 400}}
		& \multirow{1}{*}{\parbox{3.1cm}{\centering Temporal ensembling\cite{laine2016temporal}}}
		& \multirow{1}{*}{\parbox{2cm}{\centering 0.762 $\pm$ 0.049}} 
		\\
		&
		& \multirow{1}{*}{\parbox{3cm}{\centering Mean teacher\cite{tarvainen2017mean}}}
		& \multirow{1}{*}{\parbox{2cm}{\centering 0.795 $\pm$ 0.033}} 
		\\
		&
		& \multirow{1}{*}{\parbox{3cm}{\centering Dense MI-VAT}}
		& \multirow{1}{*}{\parbox{2cm}{\centering 0.858 $\pm$ 0.008}} 
		\\
		
		& 
		& \multirow{1}{*}{\parbox{3cm}{\centering Sparse-Uniform MI-VAT}}
		& \multirow{1}{*}{\parbox{2cm}{\centering 0.833 $\pm$ 0.031}} 
		\\
		
		& 
		& \multirow{1}{*}{\parbox{3cm}{\centering Sparse-Attention MI-VAT}}
		& \multirow{1}{*}{\parbox{2cm}{\centering \textbf{0.869} $\pm$ 0.021}} 
		\\
		\cmidrule{2-4}
		& \multirow{5}{*}{\parbox{1cm}{\centering 800}}
		& \multirow{1}{*}{\parbox{3.1cm}{\centering Temporal ensembling\cite{laine2016temporal}}}
		& \multirow{1}{*}{\parbox{2cm}{\centering 0.743 $\pm$ 0.037}} 
		\\
		&
		& \multirow{1}{*}{\parbox{3cm}{\centering Mean teacher\cite{tarvainen2017mean}}}
		& \multirow{1}{*}{\parbox{2cm}{\centering 0.777 $\pm$ 0.026}} 
		\\
		&
		& \multirow{1}{*}{\parbox{3cm}{\centering Dense MI-VAT}}
		& \multirow{1}{*}{\parbox{2cm}{\centering 0.865 $\pm$ 0.021}} 
		\\
		
		& 
		& \multirow{1}{*}{\parbox{3cm}{\centering Sparse-Uniform MI-VAT}}
		& \multirow{1}{*}{\parbox{2cm}{\centering 0.874 $\pm$ 0.027}} 
		\\
		
		& 
		& \multirow{1}{*}{\parbox{3cm}{\centering Sparse-Attention MI-VAT}}
		& \multirow{1}{*}{\parbox{2cm}{\centering \textbf{0.902} $\pm$ 0.019}} 
		\\
		\midrule[0.7pt]
		\multirow{11}{*}{\parbox{1cm}{\centering 400}} 
		& \multirow{1}{*}{\parbox{1cm}{\centering 0}}
		& \multirow{1}{*}{\parbox{2cm}{\centering Baseline}}
		& \multirow{1}{*}{\parbox{2cm}{\centering 0.860 $\pm$ 0.029}} 
		\\\cmidrule{2-4}
		
		& \multirow{5}{*}{\parbox{1cm}{\centering 400}}
		& \multirow{1}{*}{\parbox{3.1cm}{\centering Temporal ensembling\cite{laine2016temporal}}}
		& \multirow{1}{*}{\parbox{2cm}{\centering 0.769 $\pm$ 0.054}} 
		\\
		&
		& \multirow{1}{*}{\parbox{3cm}{\centering Mean teacher\cite{tarvainen2017mean}}}
		& \multirow{1}{*}{\parbox{2cm}{\centering 0.851 $\pm$ 0.020}} 
		\\
		&
		& \multirow{1}{*}{\parbox{3cm}{\centering Dense MI-VAT}}
		& \multirow{1}{*}{\parbox{2cm}{\centering 0.919 $\pm$ 0.008}} 
		\\
		
		& 
		& \multirow{1}{*}{\parbox{3cm}{\centering Sparse-Uniform MI-VAT}}
		& \multirow{1}{*}{\parbox{2cm}{\centering \textbf{0.920} $\pm$ 0.016}} 
		\\
		
		& 
		& \multirow{1}{*}{\parbox{3cm}{\centering Sparse-Attention MI-VAT}}
		& \multirow{1}{*}{\parbox{2cm}{\centering 0.917 $\pm$ 0.014}} 
		\\
		\cmidrule{2-4}
		& \multirow{5}{*}{\parbox{1cm}{\centering 800}}
		& \multirow{1}{*}{\parbox{3.1cm}{\centering Temporal ensembling\cite{laine2016temporal}}}
		& \multirow{1}{*}{\parbox{2cm}{\centering 0.759 $\pm$ 0.030}} 
		\\
		&
		& \multirow{1}{*}{\parbox{3cm}{\centering Mean teacher\cite{tarvainen2017mean}}}
		& \multirow{1}{*}{\parbox{2cm}{\centering 0.856 $\pm$ 0.017}} 
		\\
		&
		& \multirow{1}{*}{\parbox{3cm}{\centering Dense MI-VAT}}
		& \multirow{1}{*}{\parbox{2cm}{\centering 0.928 $\pm$ 0.012}} 
		\\
		
		& 
		& \multirow{1}{*}{\parbox{3cm}{\centering Sparse-Uniform MI-VAT}}
		& \multirow{1}{*}{\parbox{2cm}{\centering 0.919 $\pm$ 0.011}} 
		\\
		
		& 
		& \multirow{1}{*}{\parbox{3cm}{\centering Sparse-Attention MI-VAT}}
		& \multirow{1}{*}{\parbox{2cm}{\centering \textbf{0.928} $\pm$ 0.006}} 
		\\
		\bottomrule
	\end{tabular}
      }
	\label{tab:results_cifar10}
\end{table}

\begin{table*}[!h]
	\centering
	\caption{Basic demographic characteristics of the labelled and unlabelled
		cohorts used in the tremor detection experiment.
                Provided values are the mean of the population with the standard
                deviation in parentheses.
		PD status for the unlabelled data was self-reported by the participants
		themselves and not officially provided by neurologists. Tremor
              positive refers to whether a subject exhibits hand tremor or not (not all
            PD patients do exhibit tremor).}
	\begin{tabular}{l|ccc|ccc}
		\toprule
		& \multicolumn{3}{c}{Labelled data (introduced in
        \cite{alpapado2019tremor})} & \multicolumn{3}{c}{Unlabelled data} \\
		\midrule
                & Healthy & PD & Total & Healthy (self-reported) & PD
          (self-reported) & Total\\
		\midrule
          No. of participants & 14 & 31 & 45 & 380 & 74 & 454\\
                Age & 55.4 (11.7) & 62.1 (7.3) & 60.0 (9.4) & 54.1 (10.3) & 60.3
                (8.3) & 55.1 (10.2)\\
                Years after PD diagnosis & - & 6.3 & - & - & NA & -\\ 
                UPDRS 16 & 0.07 (0.25) & 1.09 (0.89) & - & NA & NA & - \\
                UPDRS 20 & 0 (0) & 1.19 (1.25) & - & NA & NA & -\\
                UPDRS 21 & 0 (0) & 0.96 (1.51) & - & NA & NA & -\\
                Sum of UPDRS-III & 2.28 (3.47) & 19.7 (11.5) & - & NA & NA & -\\
                Contributed acceleration segments & 441 (512) & 576 (537) & 534 (533) & 343 (347) & 278
                (319) & 333 (344)\\
		\midrule
                \textbf{Tremor positive} & 0 & 16 & 16 & NA & NA & NA \\
		\bottomrule
	\end{tabular}
	\label{tab:tremor_demographics}
\end{table*}

For MNIST, we used a LeNet-5 model for the embedding function $\phi$ (in
accordance with \cite{pmlr-v80-ilse18a}) with  an embedding of size $M$=800. The
instance embeddings were then merged via an attention mechanism with  $L$=128
and the resulting bag embedding is transformed to a class score via a single
linear layer (transformation $\rho$). For CIFAR-10, we used the Conv-Small
architecture from \cite{Miyato2016a} with $M$=192 for $\phi$, $L$=128 
and a single linear layer for $\rho$.

To gain some insight about the behaviour of our MI-VAT variants under various
conditions, we ran the experiment for multiple values of labelled bags $L$ and
unlabelled bags $U$ and compared them against i) a baseline model of the same
architecture and training details that makes no use of unlabelled data, and ii)
Mean Teacher and Temporal Ensembling, two alternative SoA SSL that can be 
directly applied on the same problem scenario by incorporating their
respectively proposed  regularization terms in the overall cost function, as
these terms depend only on the model's past predictions. Both approaches were
trained on the exact same data and tuned using the same budget for
hyperparameter tuning. For each $L$ and $U$ value, we train the model 10 times
and compute the average \emph{Area Under Receiver Operating Curve (ROC-AUC)}
across all trials.  All models were trained for $100$ epochs using the Adam
optimizer with a base learning rate of $0.001$, except for Mean Teacher that was
trained using SGD as per \cite{tarvainen2017mean}. 
The results for both experiments are presented in Tables
\ref{tab:results_mnist2} and \ref{tab:results_cifar10}. 

Based on these results, we see that MI-VAT can lead to large performance improvements, 
beating both the baseline and the alternative
SSL approaches that were considered.
More specifically, in the MNIST experiment we see improvements of up to 24 AUC
points over the baseline (Sparse-Attention - $L$=50 $U$=200), while
in the CIFAR-10 experiment we see improvements of up to 17 points
(Sparse-Attention - $L$=200 $U$=800). Regarding the 3 MI-VAT variants,
we see that Sparse-Uniform and Sparse-Attention generally result in
better performance than Dense MI-VAT, with Sparse-Attention often
resulting in the best performance overall. In addition, we also notice that the
performance improvement tends to be larger for smaller numbers of labelled bags.

\section{Improved detection of PD tremor in-the-wild} \label{sec:tremor_exp} 
Having established the validity of our approach in a controlled setting, we 
now turn our attention to the actual problem that this research effort
aspires to tackle.
First introduced in \cite{papadopoulos2019multiple} and
\cite{alpapado2019tremor}, the problem consists of analyzing hand acceleration
signals collected during the daily interaction of a person with their
smartphone, in an attempt to infer if the person exhibits hand tremor caused by
PD. To this end, a dataset of many acceleration recordings from $45$ subjects
was collected via a dedicated smartphone app that was installed on the
participants' personal smartphones. After installation, the app recorded inertial data whenever
a phone call was placed. Most importantly, data collection was carried out 
in-the-wild, that is, it ran during unscripted and free-living conditions,
without any interaction or requirement from the user, following an `install and
forget it' principle.   All $45$ participants, both diagnosed PD patients and healthy
controls, were recruited in the context of i-PROGNOSIS
project \cite{lisa_klingelhoefer_2017_1199554} and were
examined by medical experts for hand tremor and other PD symptoms. 

\begin{figure*}[!h]
	\centering
	\includegraphics[width=1\linewidth]{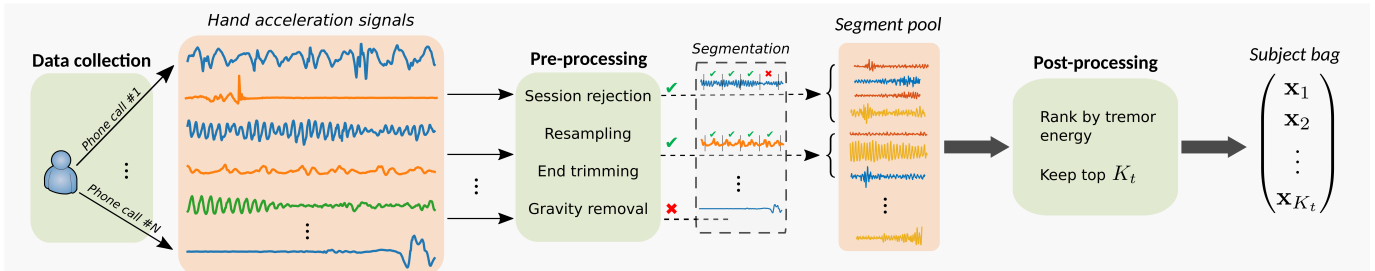}
        \caption{Schematic overview of the data collection and pre-processing
          operations that take place prior to the actual tremor detector
          training. The data of a subject, collected unobtrusively during
          free-living conditions, are transformed via a 3 stage pipeline
          (pre-processing, segmentation and post-processing stages) to a bag of
        acceleration segments.}
	\label{fig:bag_creation}
\end{figure*}

In these earlier works, the goal was to build a tremor detection model using only
data from the 45 "labelled" subjects.  As tremor is an intermittent symptom by
nature, its appearance in the bulk of the acceleration signals contributed by a
PD patient can be a relatively sparse event. Furthermore, as the data were
collected completely unobtrusively and in-the-wild, there was no way of knowing
when the tremorous episodes took place. Thus, the problem was naturally framed
as a MIL classification problem with two classes: tremor vs no tremor at the
participant level. Under this view, a participant was represented as a bag of
acceleration signal segments, accompanied by a single label that indicated
whether they exhibited tremor in general or not.  An attention-based MIL
approach was then used to train a binary classification model that resulted in
very good discrimination of tremorous vs non-tremorous participants. The
demographics of this cohort are provided in the left half of Table
\ref{tab:tremor_demographics}. The actual tremor label was created through
manual inspection of the contributed signals of a subject, in conjunction with
their neurological evaluation scores. This was necessary because using directly
the clinical scores as ground-truth can lead to  label noise in a variety of
situations  \cite{alpapado2019tremor}, caused by the unobtrusive data
collection, the intermittent/unilateral nature of tremor and the sparsity of the
neurological evaluation combined. 

To form the bag of segments encoding for a subject, a  pre-processing pipeline
was applied, during which an accelerometer signal (or \emph{session}) was
discarded if some conditions were met (short duration, low sampling frequency
$f_s$, extreme values, etc). If not, it was resampled to a common $f_s$ of 100Hz
and 5s were trimmed from both signal ends to remove picking/hanging up gestures.
Finally, a high pass filter was applied to remove the gravitational component of
the acceleration signal. The transformed signal was then segmented to
non-overlapping 5s segments. The segments from the various signals of a subject
were then ranked according to their energy in the PD tremor band (3-7Hz) and the
top $K_t$ segments were kept to form the bag of segments for a subject. 
The overall process is depicted schematically in Fig. \ref{fig:bag_creation}.

Apart from directly recruiting participants, the data collection app was also
made available in the Android app store for anyone to download and use. People
that contributed data through that channel were not examined by doctors, so
there was not any tremor ground-truth available for them.  Ultimately, this
parallel data collection  led to a second cohort of 454 people who contributed
acceleration data  but without any labels, apart from a self-reported PD status.
As a result, this larger cohort was not employed in \cite{alpapado2019tremor}
due to the lack of tremor ground-truth. The demographics of this cohort are
provided in the right half of Table \ref{tab:tremor_demographics}.

\begin{table}[!h]
	\centering
	\caption{Instance embedding transformation $\phi$
		and classification head $\rho$. $k$
		denotes the kernel size, $f$ the number of filters, 
		$s$ the stride and $M$ the embedding dimension.}
        \resizebox{\columnwidth}{!}{%
	\begin{tabular}{ccc}
		\toprule
		& $\phi$ & $\rho$ \\
		\midrule
		Input & $\mathbf{x} \in \mathcal{R}^{3\times500}$ acceleration segment & $\mathbf{z} \in
                \mathcal{R}^{64}$\\
		\midrule
		\multirow{3}{*}{Layer 1}
		& Conv1D $k=4, f=32, s=2$ & Dense $M \to 32$ \\ 
		& Leaky-ReLU ($\alpha = 0.2$) & Leaky-ReLU ($\alpha = 0.2$) \\
		& Dropout $p=0.2$ & \\
		\midrule
		\multirow{3}{*}{Layer 2}
		& Conv1D $k=4, f=64, s=2$ & Dense $32 \to 10$ \\
		& Leaky-ReLU ($\alpha = 0.2$) & Leaky-ReLU ($\alpha = 0.2$)\\
		& Dropout $p=0.2$ & \\
		\midrule
		\multirow{3}{*}{Layer 3}
		& Conv1D $k=4, f=128, s=2$ & Dense $10 \to 2$ \\
		& Leaky-ReLU ($\alpha = 0.2$) & 2-way softmax\\
		& Dropout $p=0.2$ & \\
		\midrule
		\multirow{2}{*}{Layer 4}
		& Average 1D Pooling &\\
		& Dense $128 \to 64$ & \\
		\midrule
		Output & $\mathbf{h}_k \in
		\mathcal{R}^{64}$ & $p(y|X)$\\
		\bottomrule
	\end{tabular}
      }
	\label{tab:pd_arch}
\end{table}

\begin{table}[!h]
	\centering
	\caption{Semi-supervised classification results for in-the-wild PD
          tremor detection. We report the average performance metrics across all
          LOSO iterations and random trials.
              }
	\begin{tabular}{l|c*{3}{c}}\toprule
		\textbf{Method} & \textbf{Precision} & \textbf{Specificity} &
		\textbf{Sensitivity} & \textbf{F1-score} \\\midrule
		Baseline \cite{alpapado2019tremor} & 0.854 & 0.917 & 0.875 & 0.864 \\ \midrule
		Dense & 0.938 & 0.966 & 0.938 & 0.938\\ 
		Sparse-Uniform & 0.904 & 0.945 & 0.938 & 0.920\\ 
		
		Sparse-Attention & \textbf{0.950} &
		\textbf{0.972} & \textbf{0.950} & \textbf{0.950}\\		
		\bottomrule
	\end{tabular}
	
	\label{tab:results_tremor}
\end{table}

In this paper, we wish to examine if the unlabelled data from the large cohort
can be used in a semi-supervised manner to improve the performance of a MIL
tremor detector. To this end, we follow the approach of
\cite{alpapado2019tremor} and encode each subject as a bag of acceleration
segments using the same pre-processing pipeline that was described above but we
keep the top $K$=100 segments (instead of $K$=1500 that was used in
\cite{alpapado2019tremor}) for computational efficiency purposes, as well as to
make the problem harder. After encoding each subject as a bag, we extract
features from the  raw tri-axial acceleration segments via a function $\phi$
that is a fully-convolutional 1D CNN with 4 layers, followed by average  pooling
and a fully-connected layer, that leads to an embedding  of $M$=64 dimensions.
We also use an attention dimension of $L$=128 and a fully-connected network with
3 layers for the transformation $\rho$. The model  architecture is concisely
given in Table \ref{tab:pd_arch}. We chose this particular architecture  based
on the literature and our previous experience in the field. Thus, starting from the
model of \cite{alpapado2019tremor}, which exhibited very high performance on the
tremor detection task, we introduced two modifications: replacing max-pooling
with strided convolutions and flatten/dense layers with global average pooling.
This was done in order to reduce training time without sacrificing model
capacity, as MI-VAT introduces a non-trivial computational overhead.

For our classification experiment we follow a \emph{Leave One Subject Out
(LOSO)} evaluation, in which we use the data from all labelled participants
except one to train a model, which we then evaluate on the left-out person. The
process is repeated until all participants have been left out and the results
from  each iteration are averaged.  For the baseline approach, we train the
model in a fully-supervised manner using only the labelled data, while for the
SSL approach we use both labelled and unlabelled data as per Eq.
\ref{eq:chap8:cost_function}. We do not compare against the alternative SSL
approaches here, as they did not offer significant improvement over the baseline
in our, more controlled, proof-of-concept experiments.  At each LOSO split, we
repeat model training 5 times to reduce biases caused by the random
initialization in network weights. All models were trained for $100$ epochs
using the Adam optimizer with  a base learning rate of $0.0003$. We compute the
average Precision, Sensitivity, Specificity and F1-score for a decision
threshold of $0.5$ (as per \cite{alpapado2019tremor}) across all splits and
random trials. The results of this experiment are reported in Table
\ref{tab:results_tremor}. We can see that all MI-VAT variants offer significant
improvements over the baseline, with the Sparse-Attention
variant, in particular, leading to the best performance
overall with an F1-score almost 9\% higher than the baseline model. 

\begin{figure}[!h]
    \centering
    \begin{tikzpicture}
      [every label/.append style={text=black, font=\scriptsize}]
    \begin{axis}[
    xlabel={Subjects included in the unlabelled set},
    ylabel={F1-score (\%)},
    xmin=-20, xmax=500,
    ymin=80, ymax=100,
    xtick={0,100,200,300,454},
    xticklabels={0,100,200, 300, 454},
    legend pos=south east,
    ymajorgrids=true,
    grid style=dashed,
    xmajorgrids=true,
    cycle list name=exotic,
    ]
    
    \addplot+[thick,
    error bars/.cd,
    y dir=both, y explicit,
    ]
    coordinates {
            (0, 86.4) 
            (100,87.2)
            (200,91.0)
            (300,90.5)
            (454,95.0)
    };

    \node[label={-90:{86.4}},inner sep=2pt, xshift=0.15cm] at (axis cs:0,86.4) {};
    \node[label={-90:{87.2}},inner sep=2pt, xshift=0.1cm] at (axis cs:100,87.2) {};
    \node[label={-90:{91.0}},inner sep=2pt, xshift=0.1cm] at (axis cs:200,91.0) {};
    \node[label={-90:{90.5}},inner sep=2pt, xshift=0.1cm] at (axis cs:300,90.5) {};
    \node[label={-90:{95.0}},inner sep=2pt, xshift=0.1cm] at (axis cs:454,95.0) {};

    \addplot[name path=error_top,color=green!50] 
    coordinates {
            (0, 86.4+3.5)
            (100,87.2+3.6)
            (200,91.0+3.0)
            (300,90.5+3.3)
            (454,95.0+3.3)
    };

    \addplot[name path=error_bottom,color=green!50] 
    coordinates {
            (0, 86.4-3.5)
            (100,87.2-3.6)
            (200,91.0-3.0)
            (300,90.5-3.3)
            (454,95.0-3.3)
    };

    \addplot[green!50,fill opacity=0.3] fill between[of=error_top and error_bottom];

    \end{axis}
    \end{tikzpicture}
    \caption{Tremor detection performance under different unlabelled set sizes.
      Each y-value represents the average F1-score computed over 10
      randomly-sampled subsets of the full unlabelled set, that are stratified
      with respect to the self-reported PD status of the subjects. The standard
      deviation across trials is also presented via error bars.}
    \label{fig:tremor_variable_ul}
\end{figure}
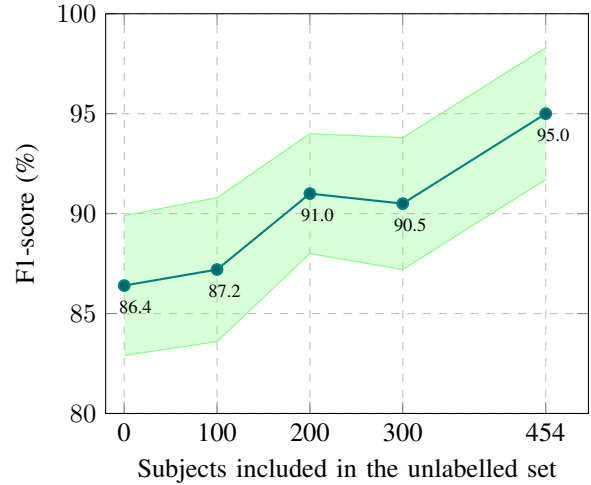

Finally, to study the effect of incorporating unlabelled data in our tremor
detection pipeline, we conduct an additional experiment, where we limit the size
of the unlabelled set to just 100, 200 and 300 subjects.  For each of these
values, we randomly sample 10 subsets from the full unlabelled set that are
stratified with respect to the self-reported PD status, to retain approximately
the composition of PD patients and Healthy subjects. Then, we train the
Sparse-Attention variant using the resulting subset and compute the average
F1-score across all 10 subsets for each unlabelled set size that we examine. The
results of this experiment are presented graphically in Fig.
\ref{fig:tremor_variable_ul}. Based on these results, it becomes evident that
there is a clear connection of increased size of unlabelled data with improved
performance, although this relation is not of a strictly increasing nature, as
exemplified by the small performance decrease between 200 and 300 subjects.

\section{Discussion}
\label{sec:discussion}
We began evaluating MI-VAT through controlled experiments on synthetic datasets.
There, we established that it can  indeed leverage unlabelled data to improve
classification performance.  An additional finding from those experiments, that
agrees with our intuition, is that performance  tends to improve when we
increase the number of unlabelled bags (Tables \ref{tab:results_mnist2},
\ref{tab:results_cifar10}). This was also the case in the real-world tremor
problem we examined later (Fig. \ref{fig:tremor_variable_ul}),  although 
it appears that the relation between performance and unlabelled set size is not
strictly increasing.
 
Stepping on these results, we then turned our attention to the real problem at
hand: using real-world unlabelled data to improve PD tremor detection
in-the-wild. To that end, we introduced a novel, unlabelled dataset of
acceleration recordings from 454 subjects that was used in conjunction with an
already available dataset of 45 labelled subjects. Through a repeated LOSO
experiment,  we showed that our MI-VAT approach leads to significant
improvements over the fully-supervised baseline of \cite{alpapado2019tremor}.
More specifically, we observed an increase in F1-score of $\sim$7.5\% for Dense,
$\sim$6\% for Sparse-Uniform and $\sim$9\% for Sparse-Attention variants, a
result that further reinforces our belief that the latter variant is the most
beneficial. The importance of successfully leveraging unlabelled data of this
type is not limited to the field of PD. In fact, it paves the way for
incorporating unlabelled data in remote, data-driven, screening pipelines for
other diseases, like Alzheimer's or depression, which also benefit from a MIL
treatment, as their symptoms can too be intermittent.

Having said that, our approach, at its current form, suffers from one important
limitation: it cannot be directly applied to any data type, due to the use of
additive noise for instance perturbation.  While an additive perturbation may be
perfectly fine for data like images or, more generally, signals that live in
vector spaces like $\mathbb{R}^N$, it may be unsuitable for others, like the
typing data of \cite{alpapado2020pd} where the bag instances
were histograms of button hold and release intervals. Using an 
additive perturbation here, would be problematic, because the perturbed vector would
cease to be a histogram (its values would no longer sum to 1).
Therefore, in such cases, modifying the perturbation operation
to ensure that the perturbed instance  does not fall
outside the data manifold is necessary.

Apart from being an interesting problem from an academic perspective, the
extension of our semi-supervised approach to additional data types is
crucial for its introduction into clinical practice of remote PD screening.
This is because patients typically exhibit only a subset of the total PD
symptomatology, thus making it difficult to accurately diagnose
the condition by looking for signs of just a single symptom.  Therefore, in
order for a remote screening tool to be successful, it must evaluate an individual for as many symptoms
of PD as possible. This can be achieved by analyzing data from a variety of
sensors that have been shown to contain diagnostic information for specific PD
symptoms. Such data sources for example, include typing patterns collected
via smartphone virtual keyboards that have been linked with fine
motor degradation, or speech recordings that may reveal signs of 
voice degradation caused by PD. Analysis of data from these sources  
could also benefit from a MIL treatment, as the corresponding PD symptom may not be
always observable especially in the early stages of the disease. In addition,
just as with the IMU-centric approach that we described here, collecting just unlabelled
data will be much easier than acquiring ground-truth by neurologists.


The proposed approach is further limited by the inherent constraints of
semi-supervised learning. For a semi-supervised method to be effective, it is
crucial that the labelled and unlabelled datasets are created in a similar manner,
that is, they are generated from the same underlying data distribution.
Unfortunately, this may not always be the case in practical scenarios. For
example, it might as well be the case that the labelled dataset is  collected
from one country, while the unlabelled dataset comes from another country with
distinct cultural practices and habits, thus leading to a shift in the data
distribution between the two datasets. This potential complication must be
thoroughly addressed during the design phase to ensure a successful real-world
deployment of a monitoring system with a semi-supervised component, like the one
proposed here.

\section{Conclusions}
\label{sec:conclusion}
Motivated by the natural coexistence of labelled and unlabelled data in remote
PD screening, we introduced a novel approach that unifies semi-supervised and
multiple-instance learning efficiently. We first established its validity 
through controlled experiments on synthetic datasets. We then went on to
significantly improve PD tremor detection in-the-wild, by leveraging IMU data
from a large cohort of unlabelled subjects. To the best of our knowledge, this
is the first work that uses unlabelled data to improve a MIL classifier in such
a way. The consistently improved performance across all experiments,
but, more importantly, its successful application on the real-world problem of
remote and unobtrusive tremor detection, 
highlights the utility of the proposed methodology.

\appendix
One issue that remains unclear is the underlying  mechanism of MI-VAT.   In
regular SSL, there is always some assumption about the data distribution that
the training procedure tries to enforce on the model's decision boundary. On a
first level, it could be argued that the observed benefit in our case, stems
from the consistency regularization effect of the MI-LDS loss term. However,
this does not directly translate to constraining the shape of the  decision
boundary, because in MIL the decision boundary  is not defined for individual
instances but for sets of instances. Hence, thinking in terms of  decision
boundary is not intuitive.

\begin{figure}[!ht]
 	\centering
 	 \begin{subfigure}{0.240\textwidth}
 	 	\includegraphics[width=\textwidth]{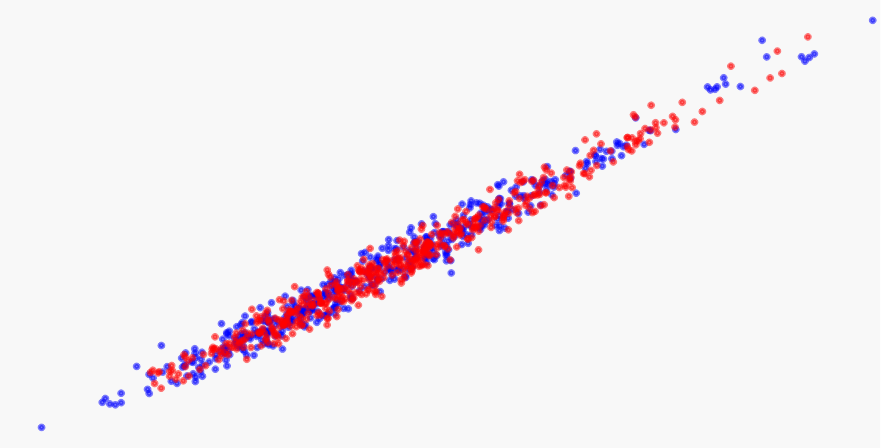}
 	 	\caption{Supervised - AUC=0.55}
 	 	\label{fig:second}
 	 \end{subfigure}
         \hfill
 	 \begin{subfigure}{0.240\textwidth}
 	 	\includegraphics[width=\textwidth]{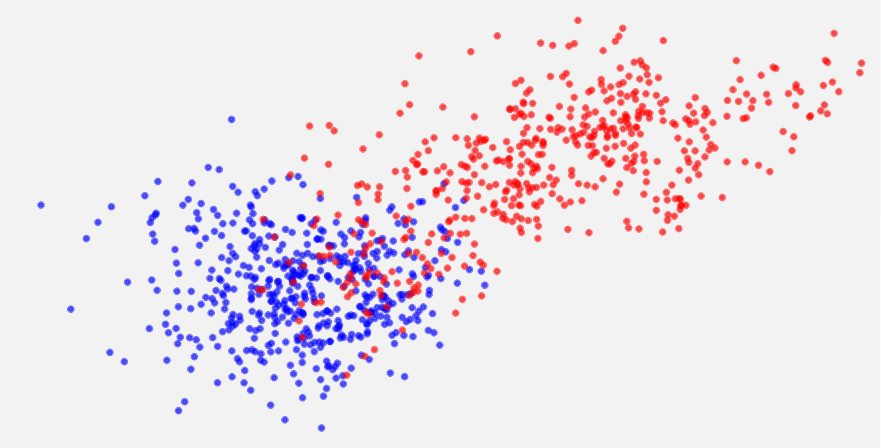}
 	 	\caption{Sparse-Attention - AUC=0.94}
 	 	\label{fig:third}
 	 \end{subfigure}
       
 	\caption{Visualization of the learned bag embeddings $\mathbf{z}$ for
 		a toy MIL problem based on the two-moons dataset.}
 	\label{fig:toy_embeddings}
 \end{figure}

 To gain some intuition about this, 
 we conduct a final exploratory experiment  based on the
 two-circles dataset, a toy 2d dataset with a spherical decision boundary. We
 sample 50 labelled, 400 unlabelled bags and  1000 test bags, without
 replacement from a two circles dataset  instantiation of 50,000 points. We then
 visualize (Fig. \ref{fig:toy_embeddings}) the learned bag embeddings of a
 fully-supervised and a Sparse-Attention model using a $\phi$ with 3
 fully-connected layers of 50, 30 and 2 hidden units. 
 We can see that the MI-LDS objective has a significant effect on
 $\mathbf{z}$,  leading to embeddings that live in two distinct clusters. On the
 contrary, using just the cross-entropy loss leads to $\mathbf{z}$ with large
 inter-class intersection.  Thus, it could be said that MI-LDS
 acts as a regularizer that encourages the inter-class bag embeddings to 
 become better separated.



\bibliographystyle{IEEEtran}
\bibliography{main}

\end{document}